\documentclass[pdflatex]{sn-jnl}
\usepackage{preambles}
\usepackage{makecell}
\usepackage{url}
\usepackage{graphicx}  % for \resizebox
\usepackage{array}
\newcolumntype{C}{>{\centering\arraybackslash}m{3.0cm}} % adjust width
\newcolumntype{L}{>{\centering\arraybackslash}m{2.2cm}} % for left header col
\usepackage{adjustbox}

\begin{document}
\title{\vspace{-15mm}MAD: Microenvironment-Aware Distillation
— A Pretraining Strategy for Virtual Spatial Omics from Microscopy\vspace{0mm}}

\author[1,2]{\large Jiashu Han}
\equalcont{\small These authors contributed equally to this work.}
\author[1,2]{\large Kunzan Liu}
\equalcont{\small These authors contributed equally to this work.}
\author[3]{\large Yeojin Kim}
\author[3,4]{\large Saurabh Sinha}
\equalocont{\small Correspondence: \href{mailto:saurabh.sinha@bme.gatech.edu}{saurabh.sinha@bme.gatech.edu} and \href{mailto:sixian@mit.edu}{sixian@mit.edu}}
\author[1,2]{\large Sixian You}
\equalocont{\small Correspondence: \href{mailto:saurabh.sinha@bme.gatech.edu}{saurabh.sinha@bme.gatech.edu} and \href{mailto:sixian@mit.edu}{sixian@mit.edu}}

\affil[1]{\small \centering{Research Laboratory of Electronics,\\Massachusetts Institute of Technology}\vspace{1mm}}
\affil[2]{\small \centering{Department of Electrical Engineering and Computer Science,\\Massachusetts Institute of Technology}\vspace{1mm}}
\affil[3]{\small \centering{The Wallace H. Coulter Department of Biomedical Engineering,\\Georgia Institute of Technology}\vspace{1mm}}
\affil[4]{\small \centering{H. Milton Stewart School of Industrial \& Systems Engineering,\\Georgia Institute of Technology}\vspace{1mm}}

\abstract{
Bridging microscopy and omics would allow us to read molecular states from images—at single-cell resolution and tissue scale—without the cost and throughput limits of omics technologies. Self-supervised pretraining offers a scalable approach with minimal labels, yet how to encode single-cell identity within tissue environments—and the extent of biological information such models can capture—remains an open question. Here, we introduce MAD (microenvironment-aware distillation), a pretraining strategy that learns cell-centric embeddings by jointly self-distilling the morphology view and the microenvironment view of the same indexed cell into a unified embedding space. Across diverse tissues and imaging modalities, MAD achieves state-of-the-art prediction performance on downstream tasks including cell subtyping, transcriptomic prediction, and bioinformatic inference. MAD even outperforms foundation models with a similar number of model parameters that have been trained on substantially larger datasets. These results demonstrate that MAD’s dual-view joint self-distillation effectively captures the complexity and diversity of cells within tissues. Together, this establishes MAD as a general tool for representation learning in microscopy, enabling virtual spatial omics and biological insights from vast microscopy datasets.
}

\maketitle
% \pagenumbering{gobble}
% \newpage
\section{Introduction}\label{sec:main}
Microscopy and omics offer two complementary windows into the complexity of a cell within its native tissue environment, and together serve as cornerstones of both basic biology and clinical pathology. Microscopy provides high-resolution, spatially continuous views of cellular morphology and microenvironmental organization, yet it lacks direct access to molecular states. Conversely, omics technologies, such as spatial transcriptomics, deliver comprehensive molecular measurements but are inherently destructive, low-throughput, and experimentally demanding. These contrasting strengths have motivated growing interest in bridging microscopy and omics: enabling molecular states to be inferred directly from routine images in a nondestructive manner that can, in principle, be extended to archived \cite{bai2024spatially,villacampa2021genome}, live \cite{horton2013vivo,you2018intravital}, or longitudinal settings \cite{ehlen2026lung,pham2025label}. Recent advances in supervised image-to-omics learning, powered by datasets that co-profile both modalities in the same tissue, have further demonstrated that routine tissue images contain sufficient cues to recover cell phenotypes \cite{zhao2024hist2cell,wu2025rosie}, infer protein localization \cite{zhang2025prediction}, and even predict aspects of gene expression variations \cite{schmauch2020deep,li2026ai,fu2025spatial,kobayashi2024prediction,he2020integrating,zeng2022spatial,comiter2023inference}, illustrating this potential by pairing rich visual features with corresponding molecular readouts.

Despite this promise, these supervised approaches face fundamental limitations that prevent their widespread adoption. They depend on scarce, expensive, and technically difficult-to-generate paired datasets where imaging and omics are perfectly co-registered \cite{lee2025spatial,zhang2023spatial,zhang2025spatial,enninful2026integration}. As a result, models trained on such data usually have poor generalization—tuned to specific tissues or stains, they often fail to generalize to new biological contexts or downstream tasks without extensive retraining and annotation \cite{li2021reinforcing,han2025system,lu2025physics,kobayashi2022self}.

This label bottleneck has motivated a shift toward self-supervised pretraining, which leverages vast, unlabeled image archives to learn meaningful representations. Advances in generic vision encoders (e.g., VAE \cite{kingma2013auto}, MAE \cite{he2022masked}, DINO \cite{caron2021emerging,oquab2023dinov2,simeoni2025dinov3}) have been adapted to microscopy and shown promise for tasks such as protein localization prediction and cell phenotyping \cite{moutakanni2025cell,murthy2025generalizable,kraus2024masked,pfaendler2023self,gupta2025subcell,chen2025visual,chen2024towards,vorontsov2024foundation,lu2024visual,ding2025multimodal,xu2024whole,aben2024towards,nakhli2024volta,filiot2023scaling}. However, these self-supervised efforts have largely diverged into two complementary paths: models that generate single-cell embeddings have been largely restricted to simplified cell-culture settings \cite{moutakanni2025cell,murthy2025generalizable,kraus2024masked,pfaendler2023self,gupta2025subcell}, which are challenging to translate to tissues with diverse cell types and complex microenvironments. In contrast, models trained on tissue images can learn representations in native tissue contexts \cite{chen2025visual,chen2024towards,vorontsov2024foundation,lu2024visual,ding2025multimodal,xu2024whole,aben2024towards,nakhli2024volta,filiot2023scaling}, but the resolution of their embeddings remains limited to the patch or slide level rather than the single-cell level.

\textcolor{black}{Most importantly, a critical gap remains: no self-supervised framework has successfully learned representations that connect visual cues to transcriptomic variation or demonstrated realistic image-to-omics prediction performance. This gap does not reflect a lack of molecular information in images; supervised models have already shown that image-derived embeddings can predict omics profiles with substantial accuracy, demonstrating that transcriptomic signals are latently encoded in the pixels \cite{schmauch2020deep,li2026ai,fu2025spatial}. Rather, the limitation lies in how representations are learned. Existing self-supervised approaches operate either at the level of isolated single cells or larger tissue patches, thereby failing to explicitly reconcile intracellular morphology with microenvironmental context within a unified single-cell embedding. We reason that cellular identity is jointly shaped by intrinsic morphological features and the surrounding microenvironment, and that integrating these complementary views during self-supervised pretraining can generate a more comprehensive representation that links to molecular-level information at single-cell resolution.}

Here, we introduce MAD (Microenvironment-Aware Distillation), a self-supervised pretraining strategy that learns single-cell embeddings directly from microscopy in tissue context. MAD is built on a vision transformer backbone and introduces dual-view joint self-distillation to learn the embedding of the same indexed cell in a unified space from two complementary, cell-centric views: a morphology view (the cell in isolation) and a microenvironment view (the cell within its surrounding neighborhood). To assess its biological validity, we evaluated MAD on multiple open-source datasets encompassing diverse tissue types, spatial resolutions, and imaging modalities. The resulting embeddings achieve state-of-the-art prediction performance on diverse downstream tasks, including cell subtyping, transcriptomic prediction, and bioinformatic inference. We also benchmarked foundation models with a similar number of model parameters on H\&E tissues and show that dual-view joint self-distillation enables MAD to outperform those foundational models pretrained on substantially larger datasets. Compared to existing supervised methods, MAD provides a data-efficient pretraining strategy that addresses the “label bottleneck,” enabling high-accuracy predictions from minimal labels. Importantly, we show that this success stems from effectively integrating morphological and contextual cues from images, allowing the model to capture the complexity and diversity of cells within tissue environments. Overall, MAD offers a robust bridge from visual phenotypes to the underlying cellular states and functions they reflect. We anticipate that this framework will serve as a generalizable paradigm for transforming microscopy images into biologically grounded single-cell representations, opening a practical path to virtual spatial omics and to deep biological insights from the vast troves of microscopy data already in hand.

\section{Results}\label{sec:results}

\subsection*{Principles of pretraining with MAD (Microenvironment-Aware Distillation)}

\textcolor{black}{The goal of pretraining on single-cell imaging data is to learn, without labels, a compact vector representation in a lower-dimensional embedding space that represents the identity of each cell.} Existing works such as CellCharter \cite{varrone2024cellcharter} and SpatialGlue \cite{long2024deciphering} in spatial omics have demonstrated the analytical power of combining single-cell information with the complexity and heterogeneity of surrounding cells. Inspired by this, we designed MAD to jointly capture cell morphology and microenvironmental context from subcellular imaging for a more complete cellular representation.

\textcolor{black}{
To achieve this, we first derive, from the tissue images (Fig. 1a), morphological and microenvironmental images for each single cell to construct the dataset. The morphological images contain only the segmented cell, enabling the model to focus on cellular organization and intracellular structure, whereas the microenvironmental images include the cell of interest together with its immediate neighborhood, allowing the model to capture contextual information about the cell’s niche and its relationships with surrounding cells and extracellular structures (see Methods for dataset preparation). 
During pretraining, we generate multiple paired global and zoomed-in local crops and feed them to the teacher and student networks, respectively, following a DINO \cite{caron2021emerging,oquab2023dinov2,simeoni2025dinov3} pretraining scheme (Fig. 1b). 
To integrate the morphology and the microenvironment views of the cell, we design a dual-view joint self-distillation loss formulated as a four-way cross-entropy objective that aligns the class tokens (learned summary embeddings in vision transformer) across the two views from the teacher and student networks. This encourages the model to learn cell-centric, context-aware, and scale-invariant representations (Fig. 1c and Supplementary Fig. S1; see Methods for network architecture and training details). 
Both networks share the same 307-million-parameter vision transformer backbone and are initialized with LVD-142M pretrained weights, providing a strong initialization that accelerates convergence and enhances representational quality.
}

\begin{figure}[!t]
\centering
\includegraphics[width=\textwidth]{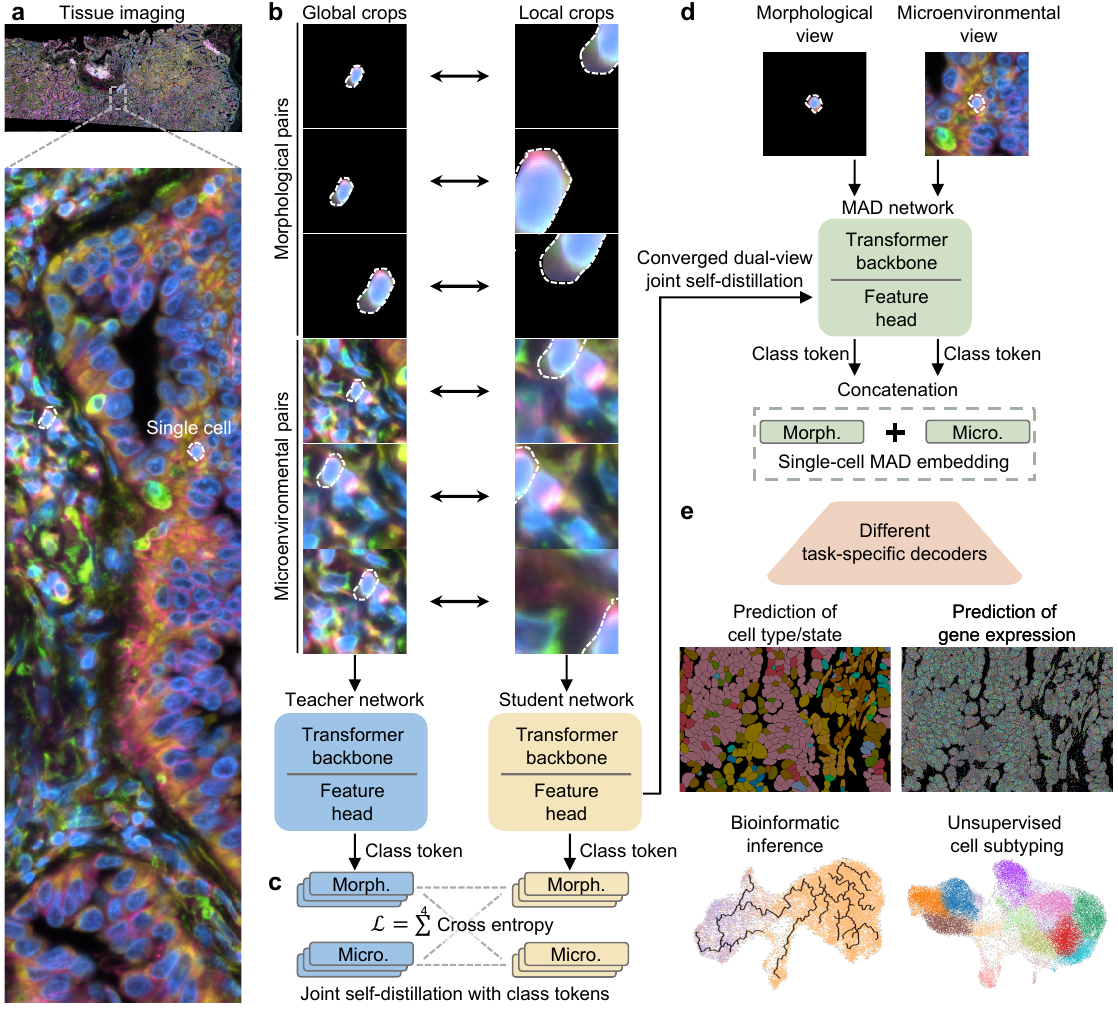}
\caption{
\textbf{Principles of pretraining with MAD (Microenvironment-Aware Distillation).}
\textbf{a}, Example tissue image with an indexed single cell highlighted; the zoom-in illustrates the cell’s subcellular morphology together with surrounding tissue context.
\textbf{b}, The dataset is prepared with morphological and microenvironmental images, from which multiple paired global and local crops are generated; the global crops are sent to the teacher network and the local crops to the student network.
\textbf{c}, \textcolor{black}{The output class tokens are aligned with a dual-view joint self-distillation objective to match the teacher (global) and student (local) representations across morphological and microenvironmental views.}
\textbf{d}, After pretraining, the single-cell MAD representation is obtained by concatenating the output class tokens from the morphological and microenvironmental views into a unified feature vector.
\textbf{e}, The resulting MAD embeddings can be integrated with task-specific decoders to support downstream analyses, including prediction of cell type/state, prediction of gene expression, bioinformatic inference, and unsupervised cell subtyping.
}
\end{figure}

Upon convergence of pretraining, the single cell embedding is obtained by concatenating the output class tokens from the target cell centered morphological and microenvironmental inputs (Fig. 1d). These embeddings can be coupled with different task specific decoders, which are substantially lighter than the pretrained MAD model, for diverse downstream prediction tasks, including cell type and state prediction, gene expression inference, bioinformatic analyses, and unsupervised cell subtyping, thereby enabling molecular level biological characterization (Fig. 1e). This establishes the foundation for pretraining with MAD on tissue images, enabling downstream applications by learning and extracting cellular complexity from both morphology and microenvironmental context.

\subsection*{MAD effectively integrates morphological and microenvironmental information to define cellular identities}

The hypothesis underlying the pretraining principles of MAD is that morphological and microenvironmental images capture distinct yet complementary aspects of cellular identity that must be integrated to fully represent a cell. Morphological images primarily encode intrinsic features such as \textcolor{black}{cell shape}, nuclear size, cytoplasmic organization, and organelle distribution, which have been shown to reflect functional state and responses to perturbation \cite{bray2016cell,seal2025cell,chandrasekaran2021image}. In contrast, microenvironmental images capture contextual cues—including spatial proximity to other cell types, extracellular matrix organization, and local tissue architecture—that modulate or even determine cell fate decisions, phenotypic plasticity, and therapeutic response \cite{engler2006matrix,de2023evolving}. Together, these morphological and microenvironmental features jointly shape cellular behavior in development and disease. For example, in ovarian cancer and other solid tumors, prior studies suggest that cellular phenotypes are strongly influenced by both intrinsic programs and contextual interactions; either source alone provides an incomplete picture \cite{hornburg2021single,yeh2024mapping}.

To test this hypothesis, we performed an ablation study of MAD’s pretraining strategies using a human ovarian cancer dataset (Fig. 2; see Methods and Supplementary Table S1 for dataset details). To evaluate the quality of different embedding strategies without omics supervision, we used the relative adjusted Rand index (ARI) in UMAP space \textcolor{black}{to quantify the segregation of different cellular phenotypes}. We found that embeddings derived from morphology or microenvironment alone captured key aspects of single-cell identity, but were less informative (reflected in lower ARI) than the integrated embeddings learned by MAD, underscoring the importance of combining both views to achieve a more complete cellular representation (Fig. 2a–c). Notably, training both image views within a single model, rather than separately in two models, was essential for integrating these complementary features into a unified embedding space (Supplementary Fig. S2). Furthermore, providing both views to the student and teacher networks—rather than assigning each network exclusively to a different view—improved training stability by maintaining consistent input distributions between the two networks (Supplementary Fig. S3). The superior embedding quality of MAD was further supported by improved training convergence and enhanced performance in downstream tasks such as cell subtyping (Fig. 2d and e; see Methods for details).

\begin{figure}[!t]
\centering
\includegraphics[width=\textwidth]{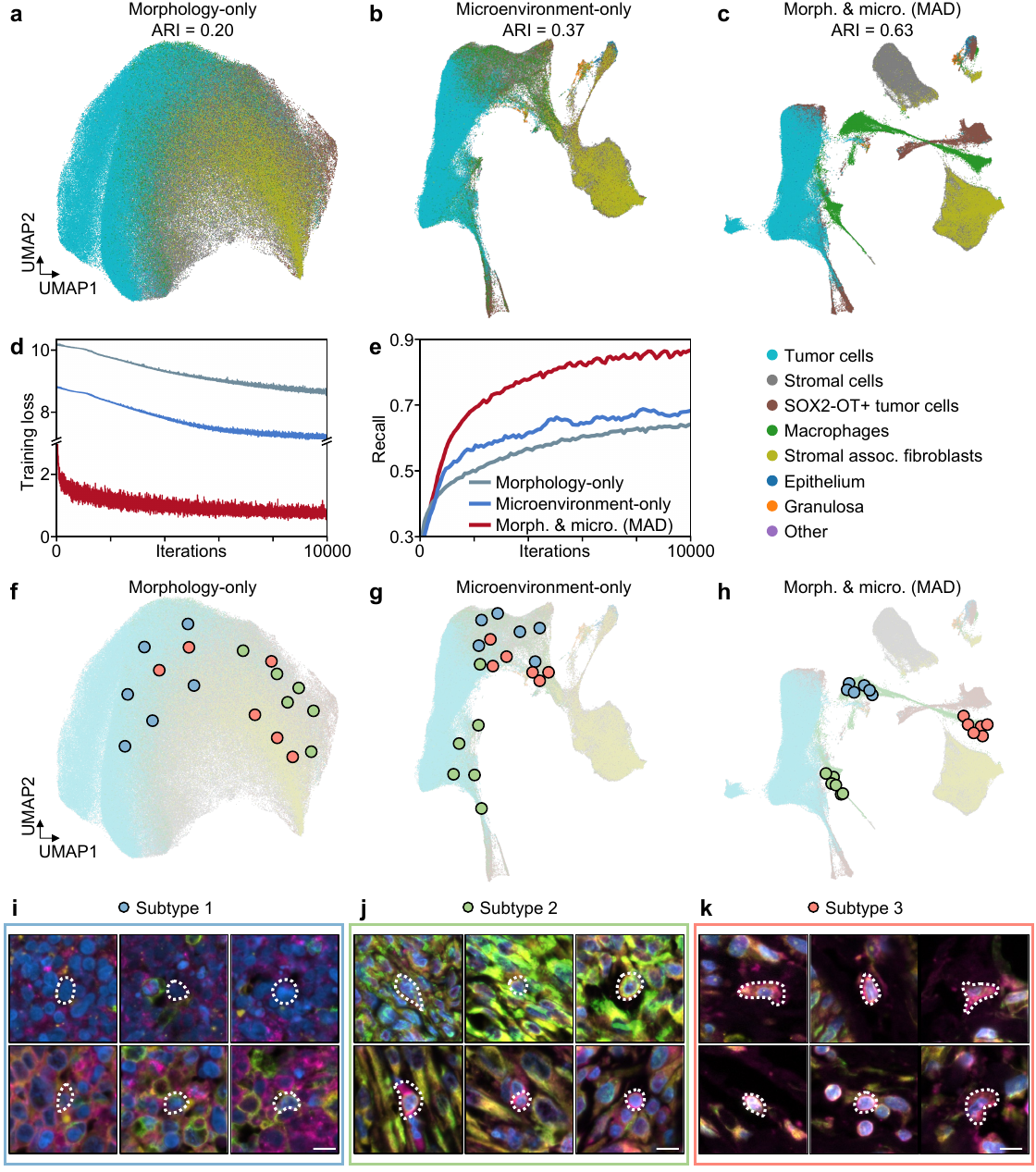}
\caption{
\textbf{MAD effectively integrates morphological and microenvironmental information to define cellular identities.}
\textbf{a}–\textbf{c}, UMAP projections of single-cell embeddings learned from morphology-only views, microenvironment-only views, and joint morphology-and-microenvironment (MAD) pretraining, respectively. Points are colored by annotated cell types; the adjusted Rand index (ARI) is shown for each embedding.
\textbf{d},\textbf{e}, Training dynamics showing self-distillation loss and recall over training iterations, respectively.
\textbf{f}–\textbf{h}, The same UMAPs as in \textbf{a}–\textbf{c}, with representative cells from three macrophage subtypes highlighted.
\textbf{i}–\textbf{k}, Example image crops corresponding to the highlighted cells for each subtype in \textbf{f}–\textbf{h}.
Scale bars: 10\,\textmu m.
}
\end{figure}

To further illustrate the biological relevance of integrating morphological and microenvironmental information, we examined macrophages as a representative example. Their diverse cellular phenotypes in both form and function make them ideal for studying these two visual domains \cite{davies2013tissue}. From the UMAP visualizations, we found that MAD, which integrates both sources of information, reveals a greater diversity of macrophage subtypes (Fig. 2a–c). Representative cells from three \textcolor{black}{known} macrophage subtypes were selected and clearly separated in the MAD embedding space, whereas they appeared more uniformly dispersed when either morphological or microenvironmental information was used alone (Fig. 2f–h). Images of these representative subtypes further highlight the complexity of cellular states: morphologically, macrophages range from rounded and amoeboid to elongated and spindle-like forms; within tissue, they occupy neighborhoods composed of different cell types such as tumor, stromal, or fibroblast cells, often in varying proportions (Fig. 2i-k). Even within the same subtype defined by omics labeling, we observed heterogeneity in both morphology and surrounding cell composition from unsupervised clustering of MAD embeddings (see Supplementary Fig. S4 for representative cells in more unsupervised subclusters). These observations emphasize that encoding both morphological and contextual information is essential to accurately recapitulate cellular diversity in biological tissues. Collectively, these ablation experiments illustrate that the performance gains of MAD arise from dual-view joint self-distillation, which effectively integrates morphological and microenvironmental features to define cellular identities, resulting in a more comprehensive embedding space.

\subsection*{Benchmarking MAD on Cell Subtyping}

\begin{figure}[!t]
\centering
\includegraphics[width=\textwidth]{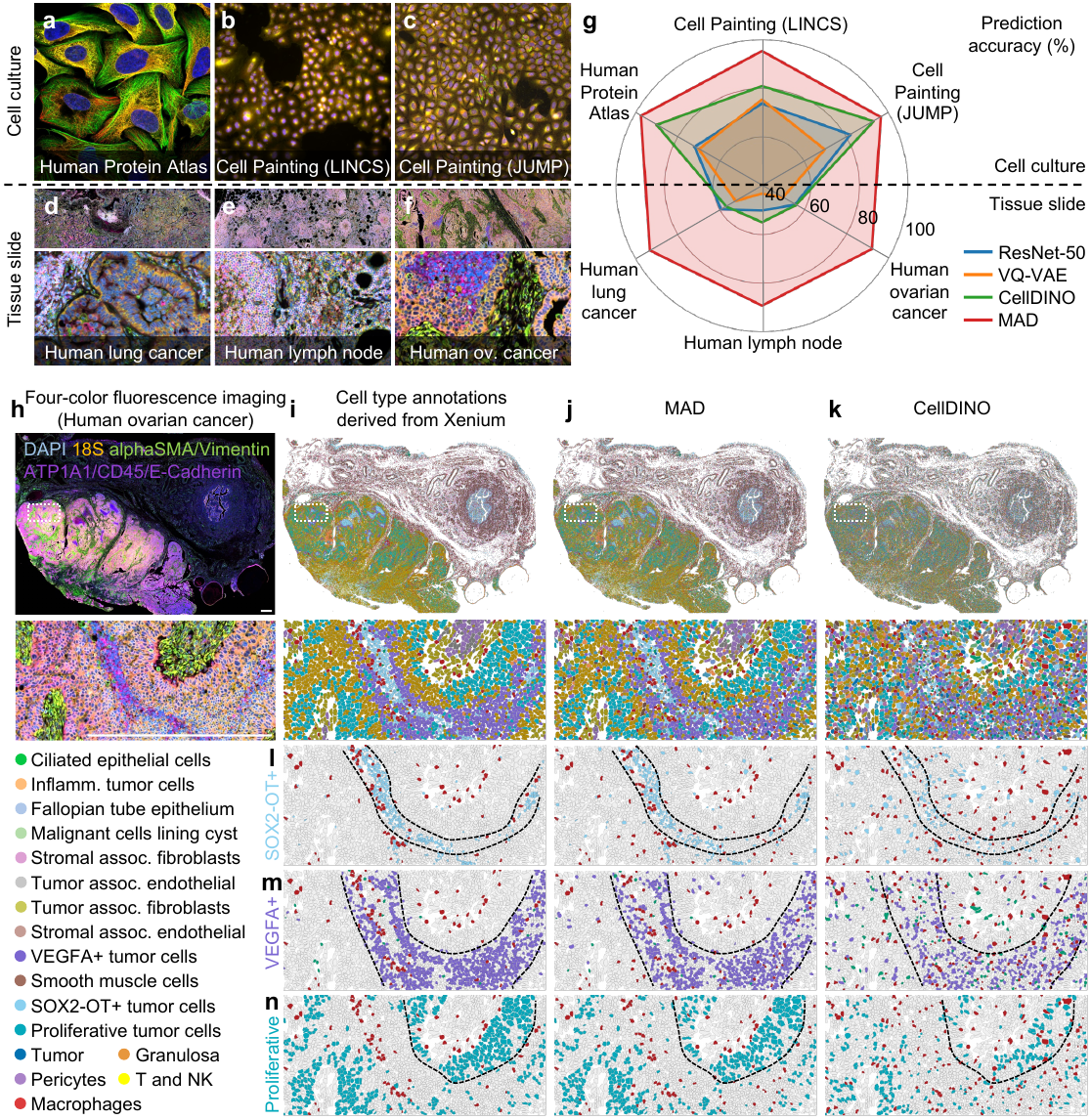}
\caption{
\textbf{Benchmarking MAD on cell subtyping.}
\textbf{a}–\textbf{f}, Representative images from six benchmark datasets spanning cell culture (top row) and tissue slides (bottom row), showing Human Protein Atlas, Cell Painting (LINCS), Cell Painting (JUMP), human lung cancer, human lymph node, and human ovarian cancer, respectively.
\textbf{g}, Summary of cell-subtype prediction accuracy across datasets (radar plot). Above dashed line: cell culture datasets.
Below dashed line: tissue slide datasets.
\textbf{h}, Four-color fluorescence imaging from the human ovarian cancer tissue dataset.
\textbf{i}–\textbf{k}, Spatial maps and corresponding zoomed-in regions of cell-type assignments for the same ovarian cancer section, showing annotations derived from Xenium, MAD, and CellDINO, respectively.
\textbf{l}–\textbf{n}, Zoomed-in comparisons of the \textcolor{black}{spatial relationships between macrophages (red) and three tumor cell subtypes}: SOX2-OT+ tumor cells, VEGFA+ tumor cells, and proliferative tumor cells, respectively.
Scale bars: 500\,\textmu m.
}
\end{figure}

To showcase the application of MAD pretraining in downstream biological tasks, we first benchmarked its performance on the task of cell subtyping. Six datasets were selected for this task, including three cell culture datasets from the Human Protein Atlas ($\sim$70,000 cells, 8 classes), Cell Painting (LINCS) ($\sim$1,000,000 cells, $\sim$100 classes), and Cell Painting (JUMP) ($\sim$1,000,000 cells, 9 classes), as well as three tissue datasets from human lung cancer ($\sim$160,000 cells, 24 classes), human lymph node ($\sim$700,000 cells, 28 classes), and human ovarian cancer ($\sim$400,000 cells, 18 classes) (Fig. 3a–f; see Methods and Supplementary Table S1 for dataset details). All datasets are multi-color fluorescence images staining different cellular architectures (e.g., nucleus, organelles, cell membrane, etc.; staining targets vary across datasets) with paired cell subtype labels. The three cell culture datasets are commonly used for subtype prediction, whereas the tissue datasets present greater challenges due to their more heterogeneous environments and a larger diversity of cell types. To demonstrate the performance of MAD embeddings for cell subtyping, we benchmarked them against three other methods: ResNet-50 \cite{he2016deep}, a standard supervised model used as a classification baseline; and VAE \cite{kingma2013auto} and CellDINO \cite{moutakanni2025cell}, two representative self-supervised image embedding methods from computer vision that also learn embeddings from individual images (Fig. 3g; see Methods for comparison details).

Our results show that MAD achieves superior performance across all datasets, reaching multi-class classification accuracy exceeding 80\% using only a two-layer multi-layer perceptron as the downstream decoder (Fig. 3g; see Methods for decoder details). This improvement is particularly pronounced in tissue datasets compared to cell-culture datasets, where the microenvironment is more heterogeneous and complex, making the incorporation of contextual information especially critical. The enhanced performance is further reflected in the clearer separation of different cell types in the embedding space, as the inclusion of microenvironmental cues in tissue datasets leads to improved cell-type distinction in UMAP visualizations (Supplementary Fig. S5). \textcolor{black}{With its high classification accuracy, pretraining with MAD preserves the spatial organization of cell types and subtypes at single-cell resolution}, which are essential for studying cell–cell communication and cellular heterogeneity in tissue biology (Fig. 3h–n, showing a stratified tumor microenvironment and its interactions with macrophages from the human ovarian cancer dataset; see Supplementary Fig. S6 for an additional example in the human lymph node dataset). Together, these findings demonstrate that MAD learns faithful single-cell image representations with superior performance \textcolor{black}{in cell subtyping across diverse cell and tissue datasets}.

\subsection*{MAD enables prediction of gene expression profiles at single-cell resolution}

\textcolor{black}{As cell types and subtypes are known to exhibit distinct gene expression programs}, moving from phenotype-level classification to gene-level prediction enables a more mechanistic understanding of cellular identity and function. We therefore take a step further to demonstrate that MAD embeddings can predict single-cell expression profiles within tissue contexts at individual gene resolution. We continue to focus on the human ovarian cancer dataset to evaluate gene expression prediction performance across a marker panel of 126 genes, using a two-layer multi-layer perceptron built upon MAD’s cell embeddings (see Supplementary Note 1 for the complete gene panel and Methods for the decoder). \textcolor{black}{This task is substantially more challenging than cell subtyping, as it requires predicting absolute gene expression levels at single-cell resolution while accounting for the sparsity of transcriptomic data (many zero values) and gene-specific differences in dynamic range and variance (Supplementary Fig. S7).}

\begin{figure}[p]
\centering
\includegraphics[width=\textwidth]{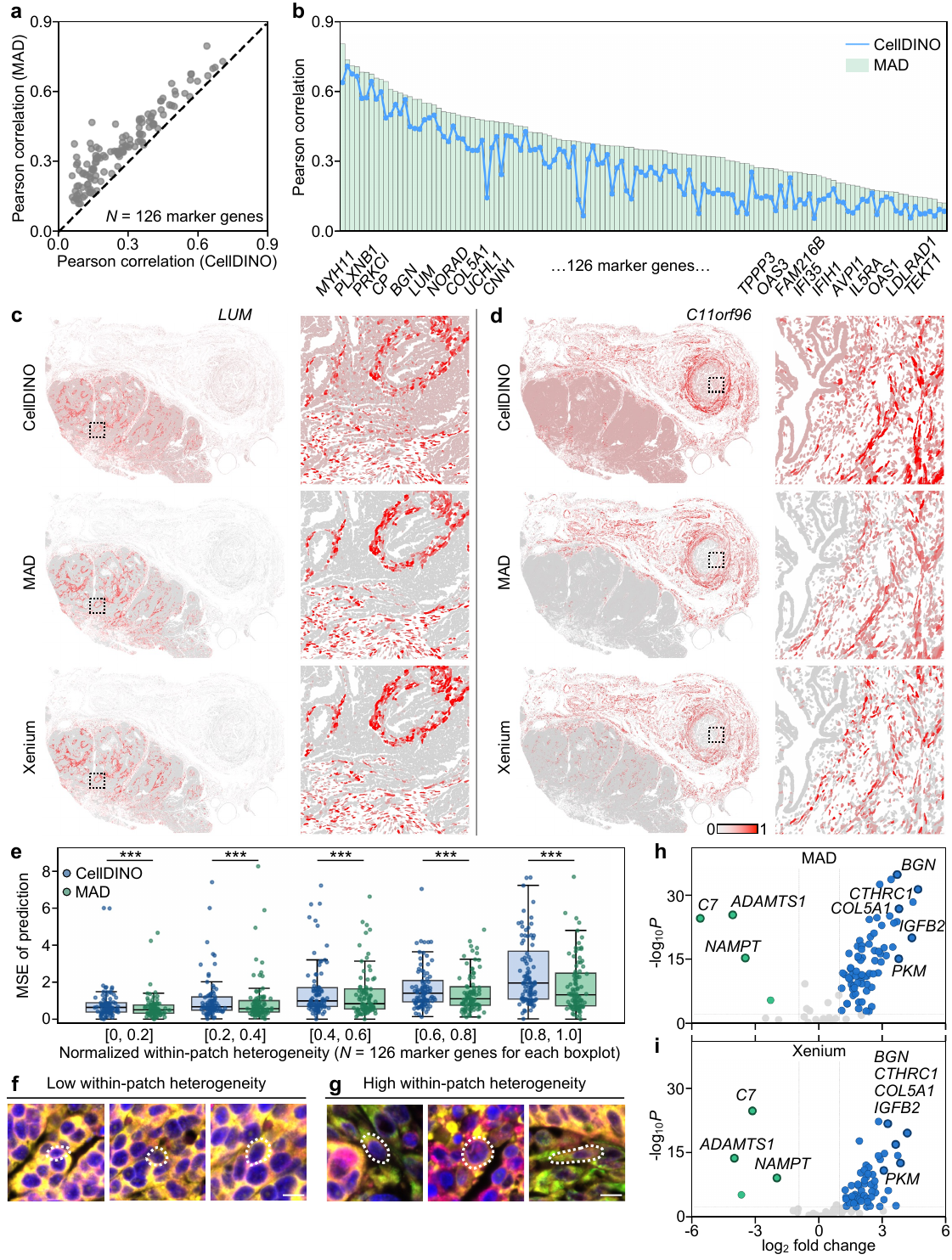}
\caption{
\textbf{MAD enables prediction of gene expression profiles at single-cell resolution.}
\textbf{a}, Per-gene Pearson correlation comparison between the benchmark method CellDINO (\textit{x}-axis) and MAD (\textit{y}-axis).
\textbf{b}, Per-gene Pearson correlation across a panel of 126 marker genes, sorted by MAD performance (highest to lowest); bars represent MAD and lines represent the benchmark method CellDINO. See Supplementary Note 1 for the complete gene panel.
\textbf{c},\textbf{d}, Spatial expression maps for \textit{LUM} and \textit{C11orf96}, shown for CellDINO, MAD, and Xenium measurements, with corresponding zoomed-in regions. The color bar indicates normalized gene expression levels.
\textbf{e}, Comparison of mean squared error (MSE) of gene expression prediction (\textit{N} = 126 genes) across different levels of within-patch heterogeneity.
\textbf{f},\textbf{g}, Representative examples of low and high within-patch gene expression variation, respectively.
\textbf{h},\textbf{i}, Comparison of differential gene expression analysis derived from MAD predictions and measurements from Xenium, respectively.
Scale bars: 10\,\textmu m.
}
\end{figure}

Compared to the baseline method CellDINO, MAD exhibits higher Pearson correlation across all 126 genes in the marker gene panel on this dataset (Fig. 4a and b; see Supplementary Note 1 for the complete gene panel). Because MAD builds on dual-view visual features that incorporate the microenvironment of the positioned cell, we found that the performance gain is especially pronounced for genes related to tissue structural features, including extracellular matrix–associated genes (\textit{COL5A1}, \textit{LUM}), muscle fiber–associated genes (\textit{MYH11}), and cell adhesion–associated genes (\textit{CDH11}) (Fig. 4b). These high-accuracy predictions, when mapped back to image coordinates, also show accurate reconstruction of gene distributions across tissues (Fig. 4c and d; see Supplementary Fig. S8 for additional examples). \textcolor{black}{Moreover, we divided the dataset into cells residing in microenvironments spanning a gradient from low to high within-patch heterogeneity in the images and evaluated gene expression prediction across all marker genes (see Methods for definitions and details). As within-patch heterogeneity increased, representing a more complex microenvironment, the performance of CellDINO decreased significantly, whereas MAD maintained consistently high predictive performance even in environments with the highest within-patch heterogeneity. These results suggest that MAD’s stronger predictive power is attributable to effective microenvironmental integration (Fig. 4e–g).} Note that the predictive ability observed here could in part stem from MAD’s strong performance in cell typing, as many of the tested genes are cell-type markers. To parse this relationship, we used a linear classifier and found that the information encoded in MAD embeddings extends beyond cell-type distinctions, suggesting that gene-level prediction is not merely a downstream consequence of accurate cell subtyping (Supplementary Fig. S9).

As a biological demonstration of the utility of single-cell gene expression prediction, we show that MAD enables exploration of differential gene expression within the tissue microenvironment. We focus on tumor- and stromal-associated fibroblasts, two critical cell populations in cancer progression and extracellular matrix remodeling (Supplementary Fig. S10a). \textcolor{black}{Differential gene expression analysis shows that MAD predictions exhibit high agreement with ground truth derived from spatial transcriptomics data and recapitulate clear molecular differences between the two cell populations (Fig. 4h and i).} For instance, \textit{BGN} (biglycan) expression is elevated in tumor-associated fibroblasts. Biglycan is a key extracellular matrix proteoglycan implicated in collagen fibrillogenesis, matrix remodeling, and pro-tumorigenic signaling. Its upregulation is associated with enhanced matrix deposition and tumor-promoting inflammation, which are hallmarks of an activated tumor stroma \cite{sorokin2010impact,gong2020damp,bonnans2014remodelling,butcher2009tense}. Spatial maps accurately reconstruct the compartmentalized distribution patterns of these genes, with significantly different expression levels between the two populations within the tumor microenvironment (Supplementary Fig. S10b). Combined with gene ontology annotations, MAD predictions also support gene set and pathway enrichment analyses due to their resolution and coverage across genes (Supplementary Fig. S11). Moreover, by leveraging spatial information preserved in MAD embeddings, we demonstrate that the model can differentiate cell subtypes based on their spatial stratification patterns in tissue images \cite{bai2024spatially} (see Supplementary Fig. S12 for an example involving macrophage subtypes). Together, these results demonstrate MAD’s ability to bridge imaging and molecular analysis, enabling a virtual measurement framework that extracts biologically meaningful patterns and holds the potential to extend to large collections of historical microscopy images.

\subsection*{Image-derived MAD embeddings recapitulate the tissue transcriptomic manifold}

\begin{figure}[!t]
\centering
\includegraphics[width=\textwidth]{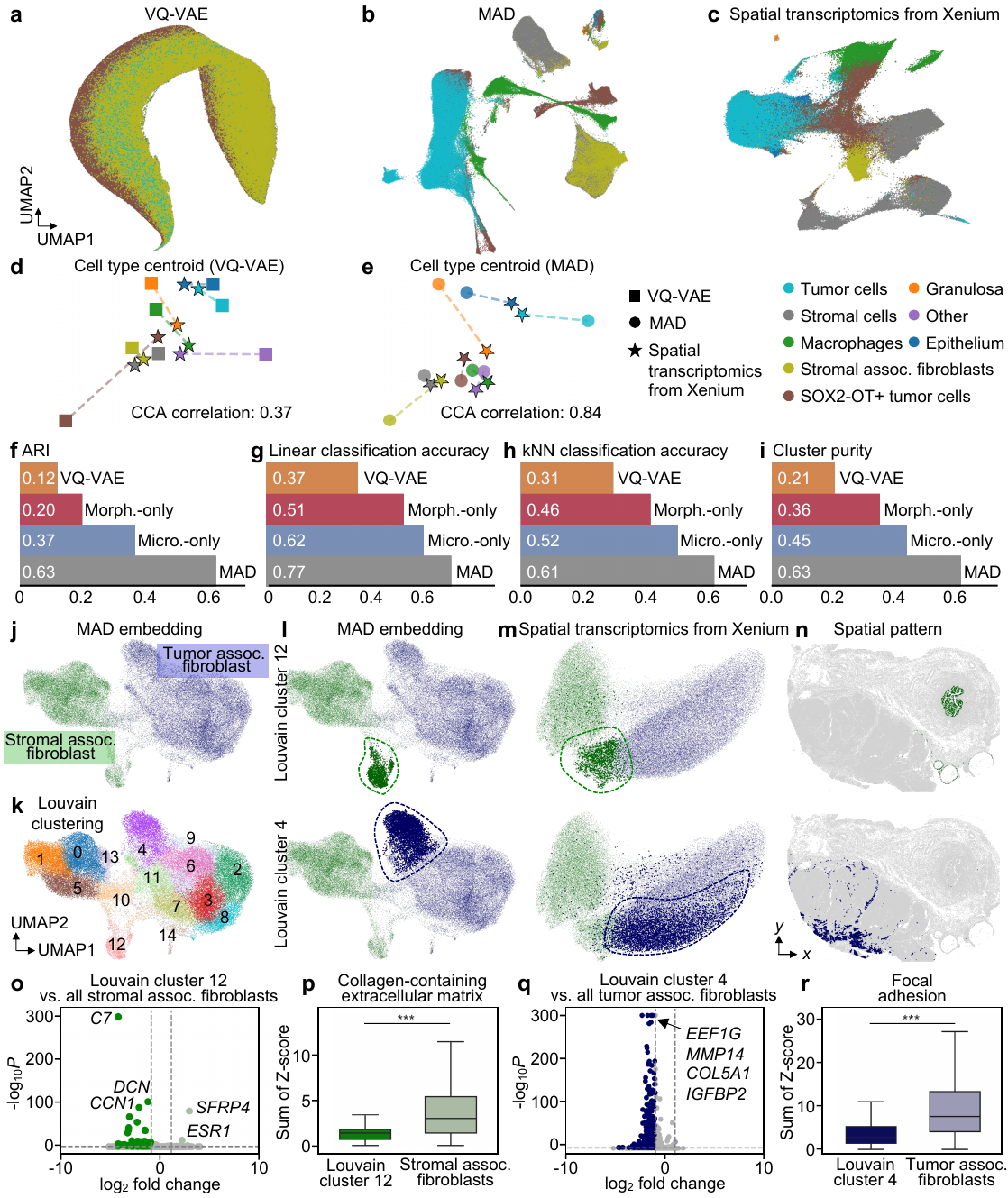}
\caption{
\textbf{Image-derived MAD embeddings recapitulate the tissue transcriptomic manifold.} 
\textbf{a}–\textbf{c}, Using the human ovarian cancer dataset, UMAPs of embeddings colored by annotated cell types from VQ-VAE, MAD, and spatial transcriptomics from Xenium, respectively.
\textbf{d},\textbf{e}, Cell-type centroid alignment between embedding and spatial transcriptomic spaces measured by canonical correlation analysis (CCA) for VQ-VAE and MAD.
\textbf{f}–\textbf{i}, Benchmark metrics comparing VQ-VAE, morphology-only, microenvironment-only, and MAD: ARI, linear classification accuracy, kNN classification accuracy, and cluster purity, respectively.
\textbf{j}, MAD embedding UMAP of fibroblasts with stromal-associated and tumor-associated fibroblasts indicated. 
\textbf{k}, Louvain clustering on the MAD embedding. 
\textbf{l}, Selected Louvain cluster 12 and cluster 4 highlighted in MAD embedding space. 
\textbf{m}, The same cells highlighted in spatial transcriptomic embedding space. 
\textbf{n}, Spatial mapping of the selected clusters back onto the tissue section. 
\textbf{o},\textbf{q}, Volcano plots for differential expression of Louvain cluster 12 vs. stromal-associated fibroblasts and Louvain cluster 4 vs. tumor-associated fibroblasts. 
\textbf{p},\textbf{r}, Gene set enrichment results for the corresponding comparisons.
}
\end{figure}

To understand how MAD enables gene-level prediction from images alone, we compared the geometric structure of its self-supervised embedding space with that of spatial transcriptomic measurements. We found that a key reason for MAD’s predictive power is that its image-derived embedding space closely recapitulates the topology of the transcriptomic manifold (Fig. 5a–c). The cross-modal similarity between MAD embeddings and spatial transcriptomics is further supported by direct alignment of the two spaces using canonical correlation analysis (CCA), which reveals a strong correspondence between their major axes of variation (Fig. 5d–e). We further show that the MAD embedding space is more informative for molecular prediction than those produced by other self-supervised strategies, as demonstrated across multiple evaluation metrics including ARI, linear classification, kNN classification, and cluster purity (Fig. 5f–i). Together, this geometric congruence indicates that MAD learns image features that align with underlying transcriptional programs. In effect, MAD constructs an image-based coordinate system that mirrors the transcriptomic landscape, providing a mechanistic basis for its ability to infer gene expression from microscopy alone.

Furthermore, this structural correspondence persists not only at the global tissue level across all cell types, but also when zooming into finer-grained substructures within a single lineage, enabling unsupervised clustering to identify cell subpopulations. We perform an analysis to tumor- and stromal-associated fibroblasts, given their well-known heterogeneity. They exhibit distinct latent representations in the MAD embedding space (Fig. 5j), and additional subclusters emerge through unsupervised Louvain clustering (Fig. 5k).

We focused on two fibroblast subclusters that are relatively segregated in latent space (clusters C12 and C4; Fig. 5l) and found that these subpopulations preserve their distinctiveness in transcriptomic space (Fig. 5m), with spatially coherent and biologically distinct localization patterns (Fig. 5n). For instance, the MAD-derived subpopulation C12 corresponds to stromal-associated fibroblasts enriched in the fallopian tube region of the ovary. Differential expression analysis between this population and the total fibroblast population (Fig. 5o) identified genes such as \textit{ESR1}, which encodes estrogen receptor $\alpha$, a key regulator of hormone-responsive epithelial and stromal biology in reproductive tissues \cite{qin2023distinct,diaz2012differential}. \textit{ESR1} overexpression in the identified subcluster is consistent with the known hormonal responsiveness and specialized extracellular matrix organization of fallopian tube stroma. Gene ontology analysis further revealed significant enrichment in collagen-containing extracellular matrix deposition and remodeling, processes characteristic of fallopian tube structure and function (Fig. 5p).

In a second example, the MAD-derived subcluster C4 identifies a morphologically distinct subpopulation of tumor-associated fibroblasts localized to the tumor periphery (Fig. 5n). It is also relatively segregated in the transcriptomic space, suggesting a distinct phenotype from the parent population of tumor-associated fibroblasts (Fig. 5m). Differential expression analysis between this group and the remaining tumor-associated fibroblasts revealed multiple downregulated genes, including canonical cancer-associated fibroblast markers such as \textit{FAP}, \textit{POSTN}, \textit{TNC}, \textit{PALLD}, and \textit{MME}, as well as ECM-related genes including \textit{COL5A1}, \textit{MMP14}, and \textit{IGFBP2} \cite{zhao2023new} (Fig. 5q). Notably, these genes are significantly enriched for the gene ontology term “focal adhesion” (Fig. 5r). The aggregated expression of this pathway is reduced in subcluster C4 compared to other tumor-associated fibroblasts, consistent with a boundary-associated, partially activated cell–matrix interaction program previously reported at tumor–stromal interfaces \cite{liu2025conserved}. Given its spatial localization and transcriptional profile, we hypothesize that this subpopulation represents a distinct stage of tumor progression relative to the tumor core.

We also show that this mirrored transcriptomic manifold supports bioinformatic inference directly in the MAD embedding space, for example by inferring pseudotime trajectories \cite{saelens2019comparison} from image-derived representations (Supplementary Fig. S13a). The trajectories inferred from MAD embeddings exhibit sharp transitions between cell subtypes, indicating their utility in identifying biologically meaningful state changes (Supplementary Fig. S13b). We validated these predictions against spatial transcriptomics–derived trajectories and observed strong concordance between image-derived and molecular trajectories at the single-cell level (Supplementary Fig. S13c). These results demonstrate that MAD embeddings derived from microscopy images preserve global transcriptomic structure while also capturing continuous developmental or state transitions.

\subsection*{Applications in spatial transcriptomics prediction from standardized histology stains}

\begin{figure}[!t]
\centering
\includegraphics[width=\textwidth]{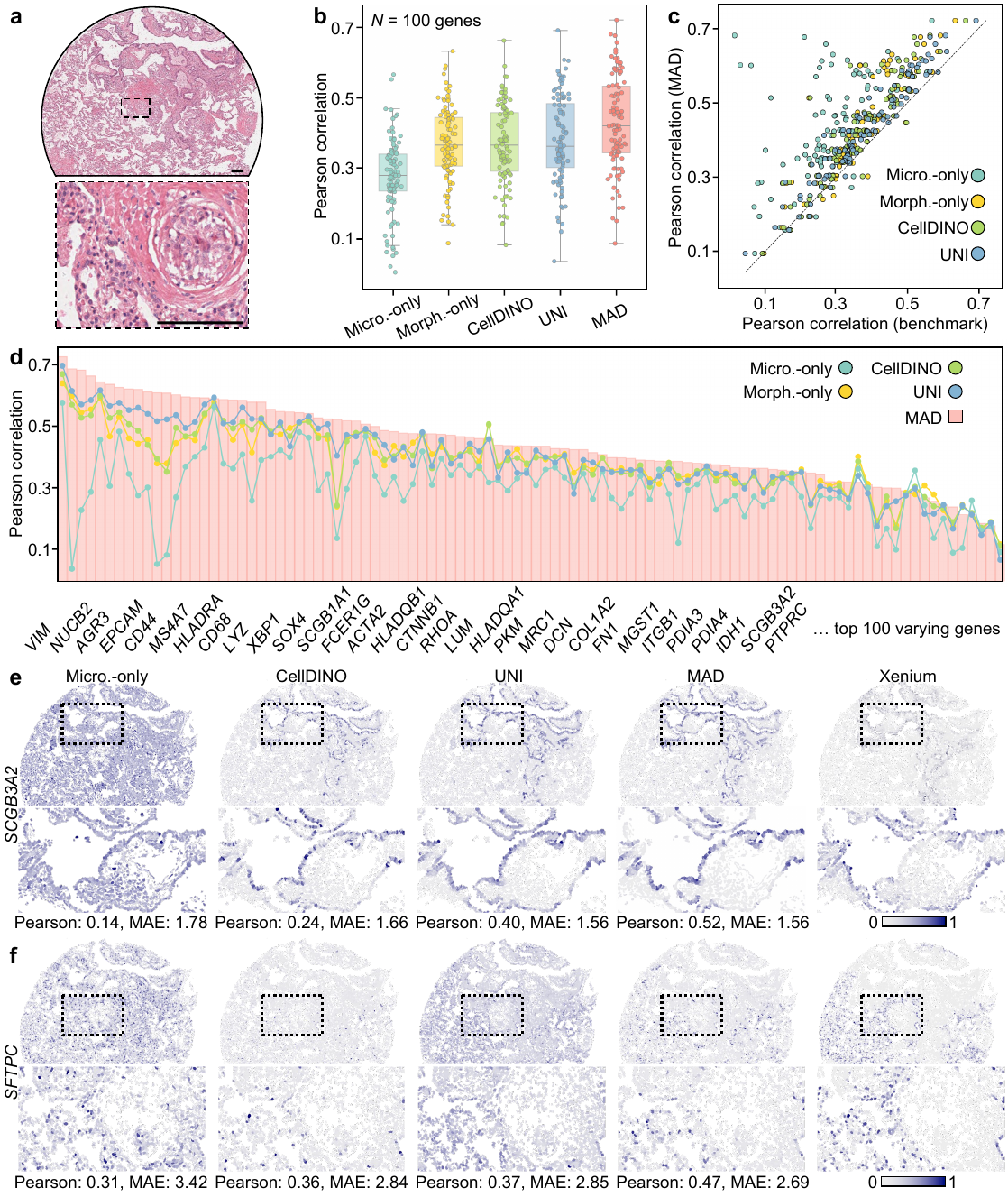}
\caption{
\textbf{Applications in spatial transcriptomics prediction from standardized histology stains.}
\textbf{a}, Representative H\&E whole-slide image used for evaluation, with a zoomed-in region illustrating cellular morphology and extracellular organizations.
\textbf{b}, Distribution of per-gene Pearson correlations for five models (microenvironment-only, morphology-only, CellDINO, UNI, MAD).
\textbf{c}, Per-gene Pearson correlation comparison between each benchmark method (\textit{x}-axis) and MAD (\textit{y}-axis).
\textbf{d}, Per-gene Pearson correlation across the top 100 varying genes, sorted by MAD performance (highest to lowest); bars indicate MAD, and lines indicate benchmark methods (microenvironment-only, morphology-only, CellDINO, UNI). See Supplementary Note 2 for the complete gene panel.
\textbf{e},\textbf{f}, Spatial expression prediction of \textit{SCGB3A2} and \textit{SFTPC} using microenvironment-only, CellDINO, UNI, and MAD, compared with Xenium measurements, with corresponding zoomed-in regions. The color bar indicates normalized gene expression levels. Pearson correlation (higher is better) and mean absolute error (MAE; lower is better) are reported for each prediction.
Scale bars: 200\,\textmu m.
}
\end{figure}

Predicting transcriptomic profiles from standardized histology stains such as Hematoxylin and Eosin (H\&E) has attracted substantial interest due to its clinical and biological utility. H\&E is the most widely available tissue imaging modality, with extensive archives spanning diverse tissue types and patient cohorts, making it highly scalable for image-based molecular inference. Several recent studies have developed H\&E foundation models that learn image embeddings for downstream tasks such as patch-level segmentation and classification \cite{chen2025visual,chen2024towards,vorontsov2024foundation,lu2024visual,ding2025multimodal,xu2024whole}. \textcolor{black}{In this section, we show that MAD can generate more comprehensive single-cell representations from H\&E images by integrating both cellular morphology and extracellular components visible in H\&E contrast, thereby improving gene expression prediction performance.}

We build on the open-source HEST-1K dataset \cite{jaume2024hest}, which contains a large collection of paired H\&E images and spatial transcriptomics data \textcolor{black}{for gene expression prediction}. We benchmark MAD against four comparison approaches: microenvironment-only and morphology-only baselines, which incorporate either patch-level or cell-level information alone, as well as CellDINO \cite{moutakanni2025cell} and UNI \cite{chen2024towards}, two benchmark models pretrained on tissue images. For a fair comparison, all models were matched to have a similar number of model parameters, and MAD was pretrained using a subset of 65 whole-slide images with available single-cell segmentation. For performance evaluation, we focused on a lung tissue subset of HEST-1K (18 tissues with $\sim$50 million cells) in which single-cell gene expression Xenium measurement is available, and all models were fine-tuned on this dataset (Fig. 6a; see Supplementary Table S1 and Methods for dataset and evaluation details). We assessed performance using a gene panel comprising the top 100 most variable genes. \textcolor{black}{We found that for 87\% of these genes, the most accurate predictions were obtained using MAD where cell morphology and microenvironmental context were both considered}, including biologically relevant markers such as \textit{SCGB3A2} and \textit{SFTPC} that play critical roles in lung cancer–related pathways \cite{vannan2025spatial} (Fig. 6b–d; see Supplementary Note 2 for the complete gene panel). Consistent with these quantitative metrics, the improved prediction accuracy corresponded to more faithful reconstruction of spatial gene expression patterns. Because MAD predicts absolute expression levels rather than only relative trends, it more closely reflects gene expression measurements obtained from molecular assays (Fig. 6e and f; see Supplementary Fig. S14 for additional examples).

Together, these results demonstrate that MAD enables spatial gene expression prediction from H\&E images. Notably, with a comparable number of network parameters, MAD outperforms foundation models pretrained on substantially larger H\&E datasets. This highlights the strength of MAD’s pretraining strategy in learning rich, generalizable cellular representations and underscores its potential as an efficient pretraining framework for future automated microscopy mining pipelines.

\section*{Discussion}

By jointly distilling the morphology and microenvironmental views of a cell within its native tissue context, MAD learns self-supervised representations at single-cell resolution across tissue scale. Importantly, MAD eliminates the need for large, perfectly coregistered image–omics datasets during pretraining, thereby overcoming the data-scarcity bottleneck that constrains supervised image-to-omics approaches. By learning directly from abundant unlabeled microscopy images, MAD enables robust molecular inference with minimal omics supervision, substantially enhancing the scalability and practical feasibility of virtual spatial omics.

Beyond performance gains, MAD provides insight into \textcolor{black}{how a cell can be encoded from images} and how much biological information such an encoding can capture. Our results show that both the intrinsic morphology of the target cell and its surrounding microenvironment are essential components of cellular identity, and that their joint integration into a unified embedding space enables more effective single-cell representation. When learned in this manner, self-supervised embeddings encode rich molecular information for over 100 critical genes associated with cell types and subtypes. These findings provide strong evidence that appropriately designed self-supervised frameworks can recover biologically meaningful information at single-cell and molecular resolution directly from images.

As a general pretraining strategy, MAD demonstrates applicability across multiple imaging modalities, including fluorescence-based subcellular imaging and standardized histology such as H\&E. Notably, when matched in model capacity, MAD outperforms foundation models pretrained on substantially larger H\&E datasets, underscoring the importance of incorporating cell-centric and context-aware inductive biases rather than relying solely on dataset scale. Moving forward, improving generalizability across diverse tissue types, disease states, and imaging resolutions will be essential. Scaling MAD pretraining to larger and more heterogeneous microscopy archives may further enhance robustness and enable the development of universal, modality-agnostic cellular encoders. In parallel, MAD embeddings may extend beyond spatial transcriptomics to other omics modalities and generative modeling frameworks, further broadening the scope of image-derived molecular inference.

\section*{Methods} \label{sec:methods}
\subsection*{Architecture of MAD}
A high-level summary of MAD is presented in Fig. 1 with details of architectures and trainings in Supplementary Fig. S1. MAD employs a vision transformer (ViT) architecture as the image encoder to map the segmented cell-centered patch and the entire microenvironment patch (with the cell centered) into a fixed-dimensional embedding. These embeddings are then concatenated to form a unified single-cell representation. The ViT adopted in this work is a standard ViT-large (ViT-L) backbone with approximately 307 million parameters \cite{vaswani2017attention}.

Within the ViT backbone, each image $\bm X$ of size $W \times W \times C$ is first partitioned into $N$ patches of size $P \times P \times C$, where
$N = \left( W/P \right)^2$.
Each patch is flattened and linearly projected into a $D = 1024$-dimensional token:
$$
\bm z_n^{(0)} = \mathrm{Proj}\left( \mathrm{vec}(\bm x_n) \right), \quad n = 1, \dots, N.
$$
Note that the number of color channels $C$ may vary across imaging modalities, but the token dimension remains fixed at $D = 1024$ to follow the ViT-L architecture. A learnable class token $\bm z_{\mathrm{cls}}^{(0)}$ is prepended to summarize the global image representation, and a learnable positional embedding is added to preserve patch ordering:
$$
\bm Z^{(0)} = \left[ \bm z_{\mathrm{cls}}^{(0)} ; \bm z_1 ^{(0)}; \bm z_2^{(0)} ; \dots ; \bm z_N^{(0)} \right] + \bm E_{\mathrm{pos}}^{(0)}.
$$
This token sequence is then processed by $L = 24$ cascaded transformer encoder blocks with pre-normalization, multi-head self-attention (16 parallel attention heads in ViT-L), and a feed-forward multilayer perceptron.

After the final block, the output of the transformer backbone is the token sequence $\bm Z^{(L=24)}$, from which only the class token $\bm z_{\mathrm{cls}}^{(L=24)}$ is extracted as the image embedding. This class token is further processed by a lightweight feature head consisting of a 5-layer multilayer perceptron, enabling the backbone output to map representations into the space where the distillation loss is applied. After concatenating the embeddings derived from the morphological view and the microenvironmental view, the resulting MAD embedding has a dimension of $D^\prime = 2D = 2048$.

\subsection*{Open-source tissue datasets}
We used seven open-source microscopy datasets in this study. A summary of these datasets is provided in Supplementary Table 1, and all data can be downloaded from the links listed in the Data Availability section.

These include three cell culture datasets consisting of four-channel stained images: Human Protein Atlas (version 16; $\sim$70,000 cells, 8 classes), Cell Painting (LINCS) ($\sim$1,000,000 cells, $\sim$100 classes), and Cell Painting (JUMP) ($\sim$1,000,000 cells, 9 classes). In the Human Protein Atlas dataset, the four channels comprise the antibody-stained protein of interest together with three reference markers for the nucleus, microtubules, and endoplasmic reticulum. In the Cell Painting datasets, the four channels correspond to major cellular compartments, including DNA, RNA/nucleoli, ER, actin/Golgi/plasma membrane, and mitochondria. Each class represents a culture condition corresponding to a specific cell type or treatment.

In addition, we used three tissue datasets generated using Xenium technologies from 10x Genomics, each containing four-channel stains: DAPI, ATP1A1/CD45/E-cadherin, 18S, and alphaSMA/vimentin. These include human lung cancer ($\sim$160,000 cells, 24 classes), human lymph node ($\sim$700,000 cells, 28 classes), and human ovarian cancer ($\sim$400,000 cells, 18 classes). In these datasets, each class corresponds to a cell type annotated based on transcriptomic profiles with expert curation.

For H\&E imaging, we used HEST-1K, a collection of H\&E images from diverse organs and diseases. These images are in standard RGB format. HEST-1K contains 65 whole-slide images with cell segmentation masks that were used for MAD pretraining. For the downstream task of transcriptomics prediction, we focused on the lung dataset (total 20 slides) as a proof-of-concept example, incorporating single-cell transcriptomics data from Visium HD or Xenium technologies, comprising a total of $\sim$50 million cells.

All datasets have subcellular microscopic resolution. Cell segmentation masks are provided as bounded polygons delineating individual cells at pixel-level resolution. Except for the two Cell Painting datasets, segmentation masks were provided with the original data. For Human Protein Atlas, cell segmentations were generated automatically using the DPN-U-Net model specifically for the HPA dataset.

\subsection*{Pre-processing and splitting of datasets for network training and testing}

\textcolor{black}{Self-supervised pretraining is performed separately for each imaging dataset (e.g., H\&E or multi-channel fluorescence). For each dataset, the whole-field images are pre-processed according to the following protocol to make them compatible with network training.} First, for each color channel: (1) flat-field correction is performed to compensate for channel-specific illumination inhomogeneity; (2) extreme pixel intensities are clipped at the 0.01–99.99th percentiles; and (3) pixel intensities are normalized to the range 0–1. This normalization constitutes the image-level preprocessing step, in addition to the standard batch normalization applied during network training.
For H\&E images, conventional stain normalization and augmentation strategies are commonly used. However, we empirically found that the above pre-processing procedures were sufficient for the datasets used in this study.

Next, whole-field images were converted into single-cell image pairs consisting of morphological and microenvironmental views. In the morphological view, all pixels outside the target cell were masked to zero. For the microenvironmental view, we empirically determined that including approximately 20–30 neighboring cells provides a suitable balance—large enough to capture relevant spatial context, yet not so large as to introduce unrelated distant cells or excessive background.
The whole-slide images were resized from their original resolution such that each microenvironmental image had a spatial dimension of 224$\times$224 pixels. Morphological views were cropped from the corresponding microenvironmental patch and resized to 70$\times$70 pixels. These dimensions were selected based on the following considerations: (1) minimizing excessive blank (masked) regions by appropriate upsampling; (2) avoiding unnecessary upsampling to save memory; and (3) ensuring both image views can be partitioned into integer numbers of patches compatible to the ViT architecture.
Accordingly, a patch size of $P \times P = 14 \times 14$ was used. The microenvironmental images therefore contain $16^2$ patches, and the morphological images contain $5^2$ patches. Standard photometric and noise augmentations—including intensity jittering, Gaussian blur, and random flipping—were applied to improve representation invariance. All processed images were stored in H5 format to facilitate efficient training.

To prevent information leakage arising from highly correlated images or batch effects in cell culture datasets, we avoided randomly shuffling individual cells across splits. Instead, all cells from the same experimental condition were assigned exclusively to either the training or testing set, ensuring that the model does not learn batch-specific artifacts.

The dataset split consisted of three components: (1) pretraining datasets without labels; (2) fine-tuning datasets with labels for downstream supervision; and (3) testing datasets that were unseen. For six of the seven datasets (excluding H\&E), 75\% of the cells were used for pretraining and fine-tuning, and the remaining 25\% were reserved for testing. This procedure was repeated across four splits, and performance was reported as the average across the four experiments.
For the H\&E dataset, patient-level stratification was use similar to previous work \cite{jaume2024hest}. Cells from 65 whole-slide images were used for pretraining, while 18 healthy and lung cancer slides were used for fine-tuning. Lung cancer slides representing distinct subcategories were reserved for testing.

\subsection*{Network pretraining with MAD}
Training MAD follows a DINO-style student–teacher self-distillation framework and applies two views of each cell in the dataset (Fig. 1 and Supplementary Fig. S1). Convergence in self-distillation occurs when the embedding outputs from the teacher and student networks align. Denote the outputs after the transformer backbone and the feature head from both views as
\[
\begin{aligned}
& \bm{z}_{T,\mathrm{morph}}^{(i)} = h_T\left(f_T\left(\bm X_{T, \mathrm{morph}}^{(i)}\right)\right), \quad
\bm z_{S,\mathrm{morph}}^{(i)} = h_S\left(f_S\left(\bm X_{S, \mathrm{morph}}^{(i)}\right)\right), \\
&\bm z_{t,\mathrm{micro}}^{(i)} = h_T\left(f_T\left(\bm X_{T, \mathrm{micro}}^{(i)}\right)\right), \quad
\bm z_{S,\mathrm{micro}}^{(i)} = h_S\left(f_S\left(\bm X_{S, \mathrm{micro}}^{(i)}\right)\right).
\end{aligned}
\]
Here, the arguments denote the morphological and microenvironmental images, respectively. The functions $f_S(\cdot)$ and $f_T(\cdot)$ represent the student and teacher transformer backbones, while $h_S(\cdot)$ and $h_T(\cdot)$ denote the feature heads that process the class token.
To match the output embeddings while unifying the embedding space of morphology and microenvironment, a key difference between MAD and DINO lies in the loss function design. Instead of a pure cross-entropy loss between two distributions \textbf{a} and \textbf{b},
$$
\mathcal L =\mathrm{CE}(a, b) = - \sum_i a^{(i)}\cdot \log b^{(i)},
$$
MAD employs a cross-entropy objective across four distributions:
\[
\begin{aligned}
        \mathcal L_{\mathrm{MAD}} = &\mathrm{CE}(\bm p_{T,\mathrm{morph}}, \bm p_{S,\mathrm{morph}})
+ \mathrm{CE}(\bm p_{T,\mathrm{micro}}, \bm p_{S,\mathrm{micro}})\\
&+ \alpha \left(
\mathrm{CE}(\bm p_{T,\mathrm{morph}}, \bm p_{S,\mathrm{micro}})
+ \mathrm{CE}(\bm p_{T,\mathrm{micro}}, \bm p_{S,\mathrm{morph}})
\right).
\end{aligned}
\]
The probability distributions are defined as

\[
\begin{aligned}
    &\bm p_{T,\mathrm{morph}}^{(i)} = \mathrm{softmax}\left(\frac{\bm z_{T,\mathrm{morph}}^{(i)} - c}{T_T}\right),\quad 
\bm p_{S,\mathrm{morph}}^{(i)} = \mathrm{softmax}\left(\frac{\bm z_{S,\mathrm{morph}}^{(i)}}{T_S}\right),\\
&\bm p_{T,\mathrm{micro}}^{(i)} = \mathrm{softmax}\left(\frac{\bm z_{T,\mathrm{micro}}^{(i)} - c}{T_T}\right),\quad 
\bm p_{S,\mathrm{micro}}^{(i)} = \mathrm{softmax}\left(\frac{\bm z_{S,\mathrm{micro}}^{(i)}}{T_S}\right).
\end{aligned}
\]
Here, $\alpha$ is a weighting factor in the loss function, $T_T$ and $T_S$ are the temperatures of the teacher and student networks, and $c$ is the running center that prevents collapse across distributions during self-distillation.

For stability, the two networks are trained asymmetrically: no gradient updates are applied to the teacher network. Instead, the teacher parameters are updated as an exponential moving average (EMA) of the student parameters, providing a slowly evolving target.

Pretraining of MAD was performed using two A6000 GPUs with the model quantized to float16 precision for memory efficiency. Pretraining on 100 million cells requires approximately 6 hours per epoch on this hardware setup, and convergence is typically achieved after 15 epochs.

\subsection*{Downstream task-specific decoders}

To evaluate the biological and phenotypic information captured by MAD representations, we trained lightweight, task-specific decoders on top of the pretrained embeddings. The MAD embedding backbone was kept frozen, and only the decoder parameters were optimized. This strategy ensures that downstream performance reflects the intrinsic quality of the pretrained MAD representations rather than improvements arising from extensive task-specific fine-tuning.

For categorical objectives, such as single-cell subtyping and binary gene prediction, we employed a two-layer multi-layer perceptron (MLP). The hidden layer width was scaled to 2X or 4X the embedding dimension, projecting the latent features to the target classes. The classifiers were trained using the AdamW optimizer with Smooth L1 loss, and early stopping was applied based on validation set performance. To address class imbalance, we implemented class-balanced sampling or loss reweighting, depending on the specific characteristics of each dataset.

For continuous targets in gene expression prediction, we adopted a similar two-layer MLP architecture, with the hidden layer width adjusted (2X to 4X the embedding dimension) according to target complexity and label density. These models were trained by minimizing the Smooth L1 loss.

\subsection*{UMAP and unsupervised clustering}

To visualize the structure of the learned single-cell embeddings and to perform unsupervised grouping, we applied Uniform Manifold Approximation and Projection (UMAP) to the encoder outputs. For each dataset, a single embedding vector per cell was extracted from the pretrained (or frozen) backbone. Unless otherwise specified, embeddings were optionally reduced to 50 dimensions using principal component analysis (PCA) to denoise the representation and accelerate nearest-neighbor search. The reduced embeddings were then projected into two dimensions using UMAP with a cosine distance metric (Euclidean distance was used in alternative analyses). UMAP was run with standard parameters—30 neighbors, a minimum distance of 0.1, and a fixed random seed (42) to ensure reproducibility.

For unsupervised clustering, we adopted two complementary strategies depending on the analysis objective. For k-means clustering, k-means was applied directly to the cell embeddings (or PCA-reduced embeddings). The number of clusters was determined either by (i) the known number of annotated subtypes when comparing against Xenium derived ground truth, or (ii) a sweep over candidate values, with selection based on internal validation criteria such as inertia or silhouette score when labels were not used. The algorithm was run with multiple random initializations (e.g., 10 restarts), and the solution with the lowest within-cluster sum of squares was retained.

For graph-based Louvain clustering, a k-nearest-neighbor (kNN) graph was constructed in the embedding space and converted into a weighted adjacency graph using a standard similarity kernel (e.g., cosine similarity or a Gaussian kernel in Euclidean space). Louvain community detection was then performed on this graph, with a resolution parameter controlling cluster granularity (default resolution near 1.0 and tuned when necessary). This approach is particularly well suited for tissue datasets, where the embedding manifold may contain non-spherical clusters or continuous cellular trajectories.

When Xenium derived subtype labels were available, clustering performance was evaluated using standard external metrics, including adjusted Rand index (ARI), classification accuracy, and cluster purity. ARI is defined as
$$
\mathrm{ARI}=\frac{\mathrm{RI}-\mathbb{E}\{\mathrm{RI}\}}{\max\{\mathrm{RI}\}-\mathbb{E}\{\mathrm{RI}\}},
$$
where RI is the Rand index defined on pairwise agreement of assignments.
Cluster purity quantifies the fraction of samples that are assigned to the majority X label within each predicted cluster, averaged across all clusters.
It is defined as
$$
\mathrm{Purity}
=
\frac{1}{N}
\sum_{k=1}^{K}
\max_{j}
\left\vert C_k \cap L_j \right\vert,
$$
where $N$ is the total number of samples, $\{C_k\}_{k=1}^{K}$ are the predicted clusters, $\{L_j\}$ are the ground-truth classes, and $\left\vert C_k \cap L_j \right\vert$ denotes the number of samples in cluster $C_k$ that belong to class $L_j$.

\subsection*{Benchmarking methods}
We benchmarked MAD against two major families of representation learning approaches—Variational Autoencoders (VAEs) and self-supervised Vision Transformers (DINO/ViT)—to assess their utility for biological discovery. These baselines span both generative latent-variable modeling and self-distillation–based self-supervised learning. To ensure a rigorous and task-relevant comparison, all models were evaluated on two downstream tasks: cell phenotyping (Fig. 3) and gene expression prediction (Figs. 4 and 5).

To isolate representation quality from downstream model capacity, we standardized the embedding extraction procedure across methods. For all VAE-based models, CellDINO, and “cell-only” benchmarks, we extracted a single embedding per cell from a masked, cell-only crop. In contrast, for UNI and “microenvironment-only” benchmarks, embeddings were generated from patch-based inputs. MAD has an integrated embedding. DINOv2 models were used with official pretrained checkpoints as frozen feature extractors (global class token or pooled outputs). All resulting embeddings were processed using identical lightweight downstream decoders to ensure that performance differences reflected representation quality rather than decoder complexity.

For density-based latent-variable baselines, we trained a VAE to compress each crop into a continuous latent space by optimizing the evidence lower bound (ELBO), consisting of a reconstruction term and a KL divergence regularizer. The encoder posterior mean (or a sampled latent vector) was used as the per-cell embedding. We additionally implemented a VQ-VAE, which replaces the continuous latent space with a discrete codebook learned through vector quantization. The VQ-VAE was optimized using reconstruction, codebook, and commitment losses. For evaluation, we considered both the pre-quantization encoder outputs and embeddings derived from the quantized codes.

To benchmark against high-performance public encoders specialized for microscopy and pathology, we included CellDINO and UNI. CellDINO is a fluorescence-microscopy–adapted, DINOv2-based model designed for single-cell phenotyping and was evaluated following its recommended protocol. UNI is a ViT-L/16 pathology foundation encoder pretrained with DINOv2 on the Mass-100K corpus. We used the officially released weights as a frozen H\&E feature extractor and evaluated it with the same standardized downstream decoders.

\subsection*{Evaluation metrics}
To quantify the alignment between morphology and microenvironment representations, we evaluated cross-view retrieval consistency using a paired-view retrieval protocol. For each cell, one view was treated as the query embedding, and all candidate embeddings from the complementary view were ranked by cosine similarity in the shared latent space. Let $i^\star$ denote the true paired instance corresponding to query $i$. A retrieval was considered correct if the true paired embedding appeared within the top $K$ ranked candidates. We report Recall@K, defined as the fraction of queries for which the true match was retrieved among the top $K$ results, with $K \in \{1,5,10\}$ unless otherwise specified. This metric directly evaluates whether the joint distillation procedure preserves instance-level correspondence across views without relying on supervised labels.

% For predicting gene presence or absence at the single-cell level, we computed precision--recall (PR) and receiver operating characteristic (ROC) curves by sweeping the decision threshold over predicted scores. Performance was summarized using the area under the curve (AUC). Genes in the evaluation panel were ordered according to their AUC values.

For predicting absolute gene expression levels, we used the Pearson correlation coefficient to quantify linear agreement between predicted and Xenium measured expression values across cells. The Pearson correlation is defined as
$$
r = \frac{\sum_{n=1}^{N} (x_n - \bar{x})(y_n - \bar{y})}
{\sqrt{\sum_{n=1}^{N} (x_n - \bar{x})^2} \sqrt{\sum_{n=1}^{N} (y_n - \bar{y})^2}},
$$
where $x_i$ and $y_i$ denote predicted and ground-truth expression values, respectively. In addition, we report the mean absolute error (MAE) to evaluate absolute deviations in predicted expression magnitude.

To unsupervisedly assess the similarity between two embedding spaces, we applied canonical correlation analysis (CCA), which measures the maximal linear correlation between projections of two multivariate representations. Given embedding matrices $\bm X$ and $\bm Y$, CCA finds projection vectors $\bm w_x$ and $\bm w_y$ that maximize
$$
\rho = \max_{\bm w_x, \bm w_y} \mathrm{corr}(\bm X \bm w_x, \bm Y \bm w_y),
$$
thereby quantifying the shared linear structure between the two embedding spaces.

In Fig. 4e–g, we divide the dataset into a gradient of subdatasets based on their within-patch heterogeneity. The within-patch heterogeneity represents the level of complexity of the tissue microenvironment and is defined as
$$
\text{Within-patch heterogeneity} = \sum_{g=1}^G \text{Var} \left\{e_{1,g},...,e_{C,g}\right\},
$$
where $G$ denotes the size of the gene panel considered ($G=126$ in Fig. 4e–g) and $C$ is the number of cells in the patch. Here, $e_{c,g}$ represents the absolute expression level of gene $g$ in cell $c$.

\subsection*{Differential gene expression and gene ontology analysis}

To evaluate the biological utility of MAD in downstream applications, we performed differential gene expression and gene ontology (GO) enrichment analyses. Differential expression analysis was conducted between two predefined cell populations using Xenium derived subtype labels. Gene expression values were derived either from spatial transcriptomics measurements (ground truth derived from Xenium measurements) or from MAD-predicted gene expression within the evaluated gene panel. For each comparison, genes were ranked based on effect size and statistical significance, and multiple hypothesis testing correction was applied to control the false discovery rate.

For functional interpretation, GO enrichment analysis was performed on the set of significantly differentially expressed genes. Enrichment was assessed against a background gene set corresponding to the full evaluated panel using the structured Gene Ontology vocabulary, encompassing three principal domains: biological process, molecular function, and cellular component. Statistical over-representation was determined using standard enrichment testing procedures with multiple testing correction. To link functional insights back to tissue organization, significantly enriched GO terms were spatially mapped to individual cells based on their predicted or measured gene expression profiles. Aggregated expression scores for GO-defined gene sets were computed per cell and projected back to the original tissue coordinates, enabling visualization of spatial patterns of functional programs under the imaging coordinates.

\subsection*{Statistics and reproducibility}

Each experiment was repeated with five independent random seeds (including data shuffling and model initialization where applicable). Performance is reported as mean $\pm$95\% confidence intervals, computed using a 10,000-sample nonparametric bootstrap over cells (stratified by slide for tissue datasets). Model selection was performed exclusively on validation folds, and the test set was evaluated once per seed after model selection. When hyperparameters differed (e.g., in ablation studies), a fixed global search grid was defined in advance and the same selection criterion was consistently applied.

Training was implemented in PyTorch using Fully Sharded Data Parallel (FSDP) and gradient checkpointing. Mixed-precision training employed bfloat16. All experiments were conducted using the same hardware and software configurations, and all random seeds were recorded to ensure reproducibility.

All box plots follow a standard convention: box boundaries represent the first and third quartiles, the central line indicates the median, and whiskers extend to data points within 1.5$\times$ the interquartile range (IQR). Points beyond this range are plotted individually as outliers. All reported statistics were verified to yield consistent results across at least five independent random seeds.

\section*{Data availability}
All datasets used in this work are open-source. Cell culture datasets are available from: Human Protein Atlas: \href{https://v16.proteinatlas.org/cell}{https://v16.proteinatlas.org/cell} (a script provided in~\cite{zhang2025prediction} can facilitate download); Cell Painting (LINCS): \href{https://open.quiltdata.com/b/cellpainting-gallery/tree/cpg0004-lincs/}{https://open.quiltdata.com/b/cellpainting-gallery/tree/cpg0004-lincs/}; and Cell Painting (JUMPS): \href{https://open.quiltdata.com/b/cellpainting-gallery/tree/cpg0002-jump-scope/}{https://open.quiltdata.com/b/cellpainting-gallery/tree/cpg0002-jump-scope/}. Tissue slide datasets are available from: human lung cancer: \href{https://www.10xgenomics.com/datasets/preview-data-ffpe-human-lung-cancer-with-xenium-multimodal-cell-segmentation-1-standard}{https://www.10xgenomics.com/datasets/preview-data-ffpe-human-lung-cancer-with-xenium-multimodal-cell-segmentation-1-standard}; human lymph node: \href{https://www.10xgenomics.com/datasets/preview-data-xenium-prime-gene-expression}{https://www.10xgenomics.com/datasets/preview-data-xenium-prime-gene-expression}; and human ovarian cancer: \href{https://www.10xgenomics.com/datasets/xenium-prime-ffpe-human-ovarian-cancer}{https://www.10xgenomics.com/datasets/xenium-prime-ffpe-human-ovarian-cancer}. 
The H\&E dataset is available at HEST-1K: \href{https://huggingface.co/datasets/MahmoodLab/hest}{https://huggingface.co/datasets/MahmoodLab/hest}.

\section*{Code availability}
The complete implementation of MAD, along with test demos using pretrained MAD weights, is publicly available at \href{https://github.com/You-Lab-MIT/MAD}{https://github.com/You-Lab-MIT/MAD}.

\bibliographystyle{naturemag}

\bibliography{MAD}

\section*{Acknowledgements}
The work was supported by MIT startup funds, Novo Nordisk Research Development US, Inc., NSF CAREER Award (2339338), and CZI Dynamic Imaging via Chan Zuckerberg Donor Advised Fund (DAF) through the Silicon Valley Community Foundation (SVCF). J.H. acknowledges support from MIT Thomas and Sarah Kailath Fellowship. K.L. acknowledges support from the MathWorks Fellowship and the MIT Health and Life Sciences Collaborative (HEALS) Fellowship.

\section*{Author contributions}
J.H., K.L., S.S., and S.Y. conceived the idea of the project.
J.H. designed and implemented the network training and inference.
Y.K. and S.S. contributed to the analysis and interpretation in biological applications.
J.H., K.L., and Y.K. prepared the figures with inputs from S.S. and S.Y..
J.H., K.L., and S.Y. wrote the manuscript with inputs from Y.K. and S.S..
S.Y. obtained the funding and supervised the research.

\section*{Competing interests}
The authors declare no competing interests.

\end{document}